%% file: paper.tex
\documentclass[11pt,a4paper]{article}
\usepackage{times,latexsym}
\usepackage{url}
\usepackage[T1]{fontenc}
\usepackage{graphicx}
\usepackage{booktabs}
\usepackage{multirow}
\usepackage{amsmath}

\usepackage[acceptedWithA]{tacl2021v1}

\usepackage{xspace,mfirstuc,tabulary}

\newif\iftaclinstructions
\taclinstructionsfalse 
\iftaclinstructions
\renewcommand{\confidential}{}
\renewcommand{\anonsubtext}{(No author info supplied here, for consistency with
TACL-submission anonymization requirements)}

\fi

\iftaclpubformat 

\else

\fi

\title{Brain Score Tracks Shared Properties of Languages: \\ Evidence from Many Natural Languages and Structured Sequences}

\author{Jingnong Qu \and Ashvin Ranjan \and Shane Steinert-Threlkeld \\
  University of Washington \\
  \texttt{\{jingnong, ar31, shanest\}@uw.edu}
}

\usepackage{xcolor}

\usepackage{color-edits}
\addauthor[Shane]{shane}{orange}
\addauthor[Ash]{ash}{blue}
\addauthor[Jingnong]{jingnong}{teal}

\newcommand{\anonymous}[2]{\iftaclpubformat #2 \else #1 \fi}

\date{}

\begin{document}
\maketitle
\begin{abstract} 
Recent breakthroughs in language models (LMs) using neural networks have raised the question: how similar are these models' processing to human language processing?  Results using a framework called Brain Score (BS)---predicting fMRI activations during reading from LM activations---have been used to argue for a high degree of similarity.  To understand this similarity, we conduct experiments by training LMs on various types of input data and evaluate them on BS.  We find that models trained on various natural languages from many different language families have very similar BS performance. LMs trained on other structured data---the human genome, Python, and pure hierarchical structure (nested parentheses)---also perform reasonably well and close to natural languages in some cases.  These findings suggest that BS can highlight language models' ability to extract common structure across natural languages, but that the metric may not be sensitive enough to allow us to infer human-like processing from a high BS score alone. 
\end{abstract}

\section{Introduction}

Modern language models (LMs) have proven potent in imitating human use of language \citep{caiLargeLanguageModels2024, wilcoxUsingComputationalModels2024}. It is therefore an interesting question whether the language models represent languages similarly to humans. \citet{schrimpfNeuralArchitectureLanguage2021} developed Brain Score (BS) for language as a metric to quantify one important aspect of this similarity. In particular, the metric tests how well the internal representations of a language model can predict functional magnetic resonance imaging (fMRI) responses of human brains when reading text. 

While proposing the metric, \citet{schrimpfNeuralArchitectureLanguage2021} also found strong correlation between models' ability of next-word prediction and their BS performance. They used this finding as evidence that the human language understanding is also optimized for predictive processing, an interesting claim that would benefit from more careful testing. 

In this paper, we set out to test this link. If next-word prediction in language models indeed mirrors human language processing, we would expect such similarity to be language-specific. That is, if we are processing English, the prediction should be based on English instead of a typologically and structurally different language such as Indonesian. This language-specificity should then project to language models for the hypothesis to fully hold. In other words, an Indonesian language model should not perform as well as an English language model in BS evaluation with English stimuli. More details on the background of BS and the inspiration of our implementation for the testing can be found in Section~\ref{sec:rw}.

\begin{figure*}[t]
  \centering
  \includegraphics[width=0.98\textwidth]{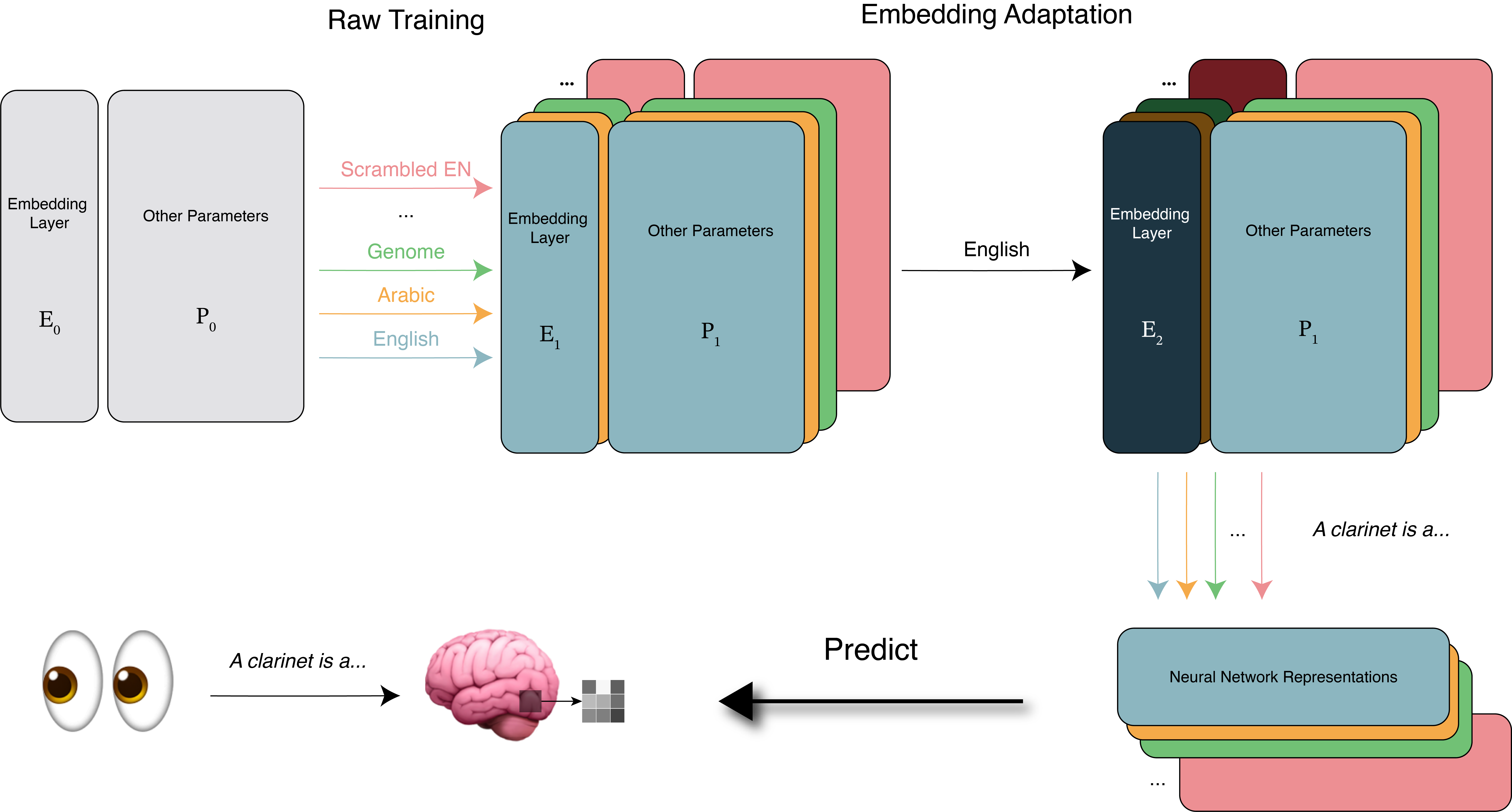}
  \label{fig:main-fig}
  \caption{The pipeline for training and evaluating the models. All training starts from a randomly initialized model. These randomly initialized models diverge in the first step by going through full training on a variety of datasets. Afterwards, they go through a separate step of embedding adaptation on English, where the embedding layers are further trained and the rest of the model is frozen. Afterwards, the model will represent English sentences used by \citet{pereiraUniversalDecoderLinguistic2018}. The sentence representations will be used to predict the human brain voxels in response to the same sentences.}
\end{figure*}

To test the language-specificity of BS, we conduct a series of experiments---depicted in Figure~\ref{fig:main-fig}---where we train a group of language models using a wide variety of natural languages and other structured sequences (the human genome, Python code, and nested parentheses).  
We then evaluate these LMs using BS on the same English reading data as above \citep{pereiraUniversalDecoderLinguistic2018,schrimpfNeuralArchitectureLanguage2021}.  In order to do this, we lightly adapt only the embedding layers of all of these models to acquire English vocabulary.
The details of the experiments are explained in Section~\ref{sec:method}.

Our results show no statistically significant difference among models trained in various natural languages with respect to their BS on two evaluation datasets. LMs trained on structured sequences have significantly higher BS performance than random baselines.  A programming language, Python, has only slightly lower BS than LMs trained on natural languages. These results, along with more detailed analyses of these experiments, may be found in Section~\ref{sec:results}.

On one hand, our findings suggest that LMs are able to extract common structure across human languages, which can be responsible for high BS scores. On the other hand, the indistinguishability between natural languages and the high scores for structured sequences also cast doubt on the hypothesized similarity between language model processing and human language processing, due to the lack of language-specificity. We discuss further ramifications of our results and avenues for refined metrics in Section~\ref{sec:discussion}.
  
\section{Related Work}
\label{sec:rw}

\subsection{Brain Score}
\label{sec:rw_bs}
\citet{SchrimpfKubilius2018BrainScore} first proposed BS as a measure of similarity between neural networks and human brains in the task of visual object recognition. \citet{schrimpfNeuralArchitectureLanguage2021} later implemented BS for natural language in several English datasets of human brain responses by comparing neural network responses to natural language stimuli and human brain imaging representations of the same responses. Here, we focus on the dataset that derives from \citet{pereiraUniversalDecoderLinguistic2018}, consistent with follow-up works to \citet{schrimpfNeuralArchitectureLanguage2021}.

Several works have since attempted to find out the contributing factors to BS for language. \citet{pasquiouNeuralLanguageModels2022} tested BS using the brain image data of participants listening to \textit{The Little Prince} and concluded that training increases models' performance in BS.

\citet{kaufLexicalSemanticContentNot2024} manipulated sentences used by \citet{pereiraUniversalDecoderLinguistic2018} as stimuli for human subjects in various ways before using them to predict the human brain activities from pretrained English language models. Computing BS of pretrained models on manipulated stimuli, they observe that manipulations that modify semantics have significantly more impact on BS than manipulations that alter syntax. They then conclude that lexical-semantic information is vital for the performance in BS, while syntactic structure is not. 

\citet{hosseiniArtificialNeuralNetwork2024} found that models training on a developmentally realistic amount of data---specifically, 100M tokens---achieve a BS nearly as high as very large models. Because of this finding, we use 100M tokens for all data types in our training procedures. 

\citet{feghhiWhatAreLarge2024} discovered that the reason for the performance of untrained GPT2-XL models, which achieve surprisingly good performance on BS, can be largely attributed to the use of shuffled train-test splits, sentence length, and sentence position. They also found that the performance of trained models in BS can be largely accounted for by sentence length, sentence position, and static word embeddings.

\subsection{Pretraining on Alternative Datasets}
\label{sec:rw_pt}
The methodology of this paper is inspired by previous works that employ alternative pretraining data in place of natural languages for neural networks. 

\citet{papadimitriouLearningMusicHelps2020,papadimitriouInjectingStructuralHints2023} have found success in pretraining on a wide variety of data, including simple formal languages, music, and typologically distinct natural languages, to lower perplexity of language models. We also base our embedding adaptation on the methods used by \citet{papadimitriouLearningMusicHelps2020} on long short-term memory (LSTM) models.

Similar procedures have also been successfully attempted and shown to be effective for different natural language downstream tasks. \citet{chiangTransferabilityPretrainedLanguage2022} tested sample manipulations on token distribution and formal languages and found success in various English downstream tasks. \citet{huCircuitsChomskyPrepretraining2025} concluded that some formal language data is more helpful than natural language data in pretraining for lowering loss and improving linguistic generalization. \citet{jiangProceduralPretrainingWarming2026} focused on procedural data, which are based on formal languages and simple algorithms, and found that front-loading such data can improve model performance on natural language, code, and informal mathematics. \citet{kimCodePretrainingImproves2024} found that pretraining on code helps model better track entities in natural languages. \citet{riPretrainingArtificialLanguage2022} created artificial languages for pretraining and discovered that a nesting dependency structure is helpful for language modeling and dependency parsing. These successes lead us to reasonably expect some level of transfer to performance in BS from training on other non-natural-language datasets.

\section{Methodology}
\label{sec:method}

Our overall methodology---depicted in Figure~\ref{fig:main-fig}---is to train language models from scratch on a variety of different datasets and then evaluate their BS score on the English reading data from \citet{pereiraUniversalDecoderLinguistic2018} after an embedding adaptation step.  We outline each component of this pipeline and detail our full experimental setup in the subsequent subsections. \anonymous{An anonymized version of the repository for the code is available at: \url{https://anonymous.4open.science/r/xlbs-CD7F}}{Code and data is available at \url{https://github.com/CLMBRs/xlbs}}.

\subsection{Datasets}
\label{sec_dataset}
We curated a group of datasets that covers training situations with different levels of similarity to English, the subject language used for calculating BS \citep{pereiraUniversalDecoderLinguistic2018, schrimpfNeuralArchitectureLanguage2021}. These datasets can be divided into three categories: natural languages, other structured sequences, and training without structures.  Exact details on corpus construction are provided in Section~\ref{sec_exp}.

\input{generated/language_family.tex}
\paragraph{Natural Languages.} To ensure a similar data quality and style across different natural languages, we use official Wikipedia dumps from November 2023 \citep{wikidumps}.
To balance typological diversity and data availability, we select 7 languages as shown in Table~\ref{tab:lang_family}. 

The experiments by \citet{pereiraUniversalDecoderLinguistic2018}, which underlies the evaluation of BS, contain two parts, Experiments 2 and 3. Experiment 2 uses only Wikipedia-style texts as stimuli for human subjects, while Experiment 3 uses both Wikipedia-style texts and first- and third-person narratives as stimuli. To have a dataset more aligned with the stimuli Experiment 3, we also include a separate dataset that combines the English Wikipedia dump with the English subset of the Project Gutenberg dataset \citep{projectGutenberg,manuProjectGutenberg} at a 3:1 ratio by example count, following a similar mix used by \citet{hosseiniArtificialNeuralNetwork2024}.  We refer to this dataset as ``Mix''.

 \paragraph{Other Structured Sequences.} We also select a variety of structured sequences that are not natural languages. These datasets include a simple formal language of nesting parentheses (the Dyck language), Python code from the Stack \citep{Kocetkov2022TheStack}, and the reference genome of Homo sapiens \citep{NCBI_Assembly_GCF00000140540_GRCh38p14}.

In Dyck language, each type of parentheses is assigned with a unique token for both opening and closing. For each token being generated, we set a probability of 0.51 where the token is a closed parenthesis, meaning the token will be the same one as the last token of an odd number in the current string, unless the token has to be open. Otherwise, it will pick a token with equal chance out of 49,999 unique tokens for the token to serve as an open parenthesis.

We preprocess the Python code by first tokenizing the code using Python's built-in tokenizer and assigning special tokens for semantically significant whitespace in Python (i.e.\ newline, indent, dedent). In addition, all comments and strings are masked with corresponding special tokens to avoid natural language leaking into the dataset.

For the human genome data, we eliminate all the headings in the dataset to again avoid natural language leakage and irrelevant information. 

\paragraph{Unstructured Training.} To set up baselines, we include a dataset of scrambled English Wikipedia. In this dataset, all tokens are scrambled across the dataset. Despite keeping some basic statistical information of the token frequencies in the dataset, the contextual dependence in natural languages is eliminated. Finally, we also test a version of the model that is not trained on any dataset after initialization.

\subsection{Brain Score}
Conceptually, BS compares the similarity between the representations of stimuli in human brains and neural networks. As shown in Figure~\ref{fig:main-fig}, sentences are represented in the language models and compared to the human fMRI images when subjects are shown the same sentences.  In particular: a linear regression model is trained to predict fMRI responses in the language network from LM activations, and BS is then the Pearson correlation between actual and predicted fMRI responses. We utilize a Python implementation of this metric from an open-source GitHub repository\footnote{\url{https://github.com/brain-score/language}} \citep{SchrimpfKubilius2018BrainScore,Schrimpf2020integrative,schrimpfNeuralArchitectureLanguage2021}. 

We compute the metric using the fMRI data from Experiments 2 and 3 conducted by \citet{pereiraUniversalDecoderLinguistic2018}. As discussed in Section~\ref{sec_dataset}, the two experiments use different styles of stimuli. A language model is evaluated on BS using this specific dataset under the method proposed by \citet{schrimpfNeuralArchitectureLanguage2021}. Each layer of a language model is treated separately. 80\% of the model representations and the human fMRI data are used for fitting a linear regression, while the remaining 20\% are used to generate the predictions. The prediction is then compared against the held-out group of the human data using a Pearson correlation. Predictivity is aggregated across all participants' voxels by calculating the median and then being divided by a ceiling value, which is the average predictivity across different human participants' data.
\begin{figure*}[ht]
  \centering
  \includegraphics[width=\textwidth]{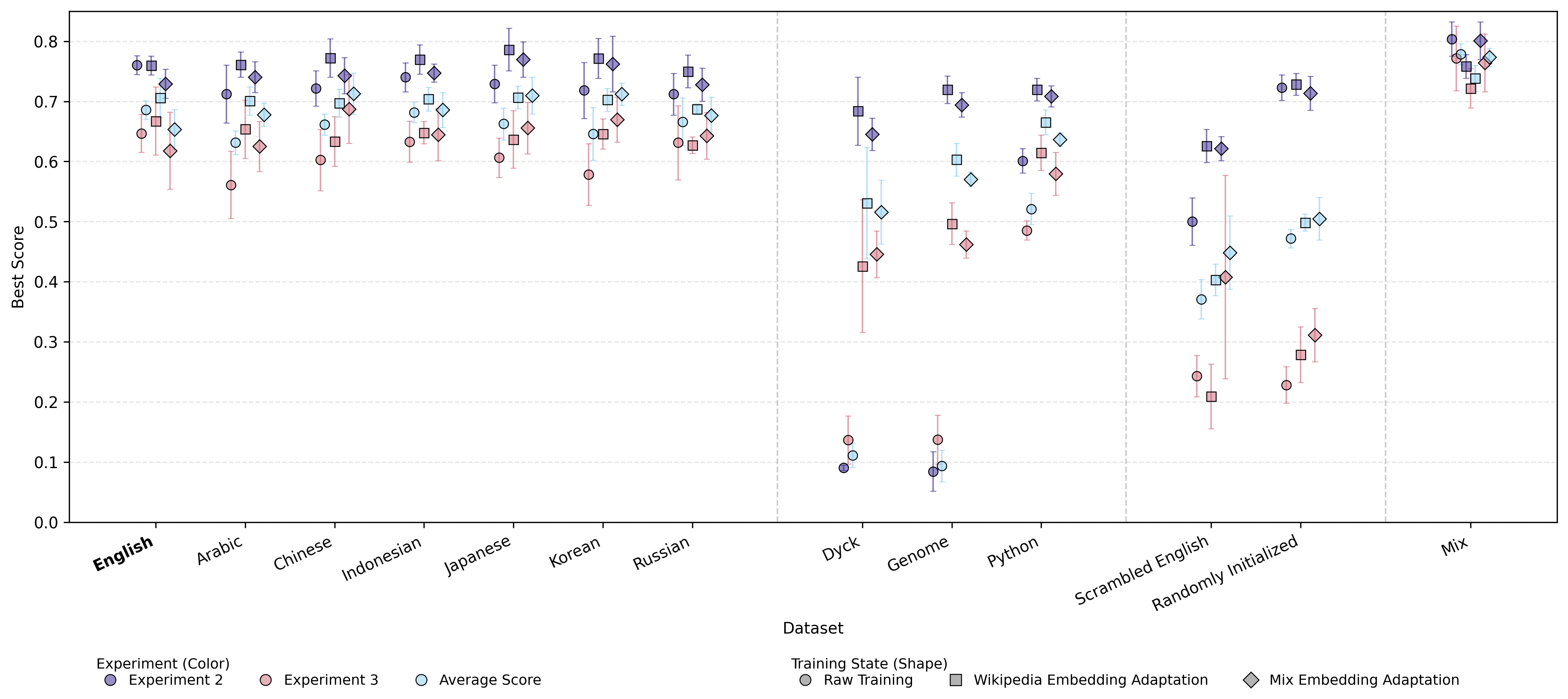}
  \caption{Effects of different embedding adaptations on BS. For each dataset in each training regime, the scores for Experiments 2 and 3 and the best average score of the two experiments are shown in different colors, with a range for the standard deviation. From left to right, each group includes scores for models trained on Wikipedia data, non-natural-language structured data, unstructured data, and mixed English data between Wikipedia and Project Gutenberg. 
  } 
  \label{fig:dataset-comparison}
  \label{fig:dataset-scatter}
\end{figure*}

\subsection{Experiments}\label{sec_exp}
Consistent with \citet{hosseiniArtificialNeuralNetwork2024}, we train a group of GPT-2-style language models with 12 layers and 117 million parameters in total \citep{radfordLanguageModelsAre2019}. The models all have a vocabulary size of 49,999 for tokens in the corpus. We conduct all trainings as well as model initialization for the untrained models on 5 different random seeds.

\paragraph{Preprocessing.}

Except for Dyck, which is tokenized with a whitespace tokenizer because it is synthetically generated, all other structured datasets are tokenized with a Byte-Pair Encoding (BPE) tokenizer. Scrambled English is scrambled after first tokenizing the English Wikipedia dataset.

\citet{hosseiniArtificialNeuralNetwork2024} report that a model trained on 100 million words can achieve predictivity similar to a fully trained model reported by \citet{schrimpfNeuralArchitectureLanguage2021}. Since all datasets have over 100 million tokens after tokenization, we truncate to 100M tokens for training for all of them. Validation splits range from 40M to 100M due to differing sizes of the original datasets (see Table~\ref{tab:valid} for full details). 
\input{generated/dataset_list_table.tex}

In order to maximize the number of tokens shown to the model we pack the datasets. This is done by joining all documents together, using the end of sequence token as a separator. After this, $k$-length slices are taken in order, where $k$ is the size of the context window, which is 1,024 for all our experiments. 

\paragraph{Raw Training.}
In the first step shown in Figure~\ref{fig:main-fig}, we train a group of randomly initialized models on next token prediction on each of the 12 types of sequences discussed in Section~\ref{sec_dataset} for 40 epochs with an early stopping mechanism with patience of 5 epochs.  
We train each model on one NVIDIA L40/L40S GPU. We use the AdamW optimizer with $\beta_1 = 0.9$, $\beta_2 = 0.999$, and $\epsilon = 10^{-8}$. The learning rate scheduler is linear with a base learning rate is $5\times10^{-5}$. Both training and evaluation batch sizes are set to 16.

\paragraph{Embedding Adaptation.} 
Because the subject language used by \citet{pereiraUniversalDecoderLinguistic2018} is English, it is important for models trained on other sequences to have at least the English vocabulary. 
Therefore, as shown in the second step in Figure~\ref{fig:main-fig}, we perform an embedding adaptation in English for all models. To do this, we further train each model with the English data while freezing all parameters except the embeddings. 
This step offers an anchor for all models to acquire light English lexical semantics. 
In line with the two types of experimental stimuli in \citet{pereiraUniversalDecoderLinguistic2018}, we perform this step separately with both English datasets, i.e. just Wikipedia or mixed with Project Gutenberg. 

\section{Results}
\label{sec:results}

Figure~\ref{fig:dataset-scatter} shows the results from the experiments described in the previous section. All scores reported in this section are the average of the scores from the best performing layers in each random seed. Qualitatively, we find very similar behaviors among natural languages across the board. There are even languages that perform better in certain tasks than the English models. At the same time, other training procedures lag slightly behind natural languages even after embedding adaptations, though Python performance on BS comes very close to natural languages. 
The highest performing models are the English models pre-trained on the mixed dataset of English Wikipedia and Project Gutenberg, which covers different domains in English and achieves close performance in the two experiments from \citet{pereiraUniversalDecoderLinguistic2018}.

Below, we report a more detailed analysis of these results. Most importantly, we confirm that all natural languages behave similarly for the BS, and Python only slightly lags behind (Section~\ref{sec:comp-model}). We also find embedding adaptation offers a moderate boost in performance in certain situations when adapting across languages (Section~\ref{sec:adapt}), but it is not effective when adapting across different domains (Section~\ref{sec:domain}).

\subsection{Effect of Training Datasets}
\label{sec:comp-model}

In this section, we focus on models that have gone through embedding adaptation on the English Wikipedia data. In Figure~\ref{fig:dataset-comparison}, by focusing on the square points, we observe that, when adapted on English Wikipedia, different natural languages exhibit very close performance in Experiments 2 and 3 from \citet{pereiraUniversalDecoderLinguistic2018}, as well as the average of the two experiments. In fact, all differences between natural language pairs are found to be statistically insignificant ($p>0.05$).\footnote{All statistical tests in this paper are Wilcoxon tests with $p=0.05$ as the threshold for statistical significance.} This is a surprising result considering the fMRI data corresponds to human brain responses to English sentences and that these models have been trained on typologically diverse languages and are only exposed to English data in the embedding layers. However, this result is also consistent with the finding by \citet{kaufLexicalSemanticContentNot2024} that lexical-semantic content mainly drives a model's performance in BS, and the syntactic differences among languages that might have been acquired by the models during training does not significantly change the outcome. 

Figure~\ref{fig:dataset-comparison} (second horizontal section) also shows that all models trained on non-natural-language data and adapted on English Wikipedia data lag slightly behind the average Brain Score of models trained on natural languages. In the most extreme example, training on scrambled English hurt the performance in Experiment 2 with a statistically significant difference compared with the randomly initialized models, even after embedding adaptation. This effect shows some form of knowledge about the structure in the data is still needed for models to perform well on BS. While trainings on Dyck, human genome, and Python offer virtually no statistically significant benefits from random initialization on Experiment 2 ($p>0.05$), they offer a statistically significant boost on Experiment 3. Specifically, we report the Hodges-Lehmann estimator ($\hat{\Delta}_{HL}$) between random initialization and Dyck, human genome, and Python to be 0.135, 0.226, and 0.343 respectively\footnote{The Hodges-Lehmann estimator is the typical difference between two groups in original scale. It is calculated as $\hat{\Delta}_{HL}=\mathrm{median}\{X_i-Y_j: i=1,..., n_1, j = 1,..., n_2\}$.}. 
Although Python models' difference from all natural languages except Russian on Experiment 2 is statistically significant with a small $\hat{\Delta}_{HL}$ ranging from $0.036$ for Arabic to $0.075$ for Japanese, the difference between the performance of the Python model and any natural language model on Experiment 3 is statistically insignificant. This finding hints that BS is similar to other evaluations or downstream tasks for language models where previous and concurrent works have found to benefit from training on various non-natural-language data \citep{chiangTransferabilityPretrainedLanguage2022,huCircuitsChomskyPrepretraining2025,jiangProceduralPretrainingWarming2026,kimCodePretrainingImproves2024,papadimitriouLearningMusicHelps2020,papadimitriouInjectingStructuralHints2023, riPretrainingArtificialLanguage2022}. Specifically, the programming language of Python is especially useful, suggesting that learning to read Python provides models with structural cues that help LM representations align with natural language reading data. 

\input{generated/combined_table.tex}

\subsection{Effect of Embedding Adaptation}
\label{sec:adapt}
We also want to know how much of the performance is solely contributed from the step of embedding adaptation. In this part, we focus on embedding adaptation using English Wikipedia. From Figure~\ref{fig:dataset-scatter} and the large percentage point in difference between raw training and embedding adaptation on Wikipedia in Table~\ref{tab:combined-table}, we can clearly see that models initially trained on Dyck, Genome, and Python receive significant improvement in both experiments after either embedding adaptation. This effect shows that the lexical semantics of English plays an important role in BS, which is also concluded by \citet{hosseiniArtificialNeuralNetwork2024}. 

For other cases, the story is not as clear. For Experiment 2, all natural language models except English and Indonesian, which show high scores prior to adaptation, demonstrate statistically significant improvement. However, for Experiment 3, only Arabic and Korean models among natural language models have shown statistically significant improvement after the same embedding adaptation. These results seem to suggest a moderate effect from embedding adaptation for natural language models, but the effect also depends on the specific language and corpus for evaluation. 

Scrambled English shows statistically significant improvement from embedding adaptation for Experiment 2 but not Experiment 3, while random initialization shows no statistically significant improvement from embedding adaptation in either experiment. 

However, the results from Experiment 2 should also be viewed with caution given the high score from a raw randomly initialized model. \citet{hosseiniArtificialNeuralNetwork2024} report that untrained models achieve an average score around 0.5 for the two experiments,\footnote{Our untrained results are effectively a replication of this fact.  \citet{hosseiniArtificialNeuralNetwork2024} also analyze which aspects of initialization drive this effect and produce an initialization scheme that does not achieve as high a BS score for untrained models.} and \citet{feghhiWhatAreLarge2024} explored in detail the contributing factors to the score of the two experiments separately for GPT2-XL. Crucially, we find that, even after embedding adaptation, statistically significant difference from the randomly initialized models without any training can only be found in Arabic and English models. The differences with Chinese, Indonesian, Japanese models are marginally insignificant ($p=0.056$).

\subsection{Effect of Different Domains}
\label{sec:domain}
Figure~\ref{fig:dataset-comparison} suggests a performance gap exists between Experiments 2 and 3 for models trained and adapted only on Wikipedia data \citep{wikidumps} (the purple and pink points in the figure). The scores of models which we train on a mixed dataset of English Wikipedia and English books in Project Gutenberg \citep{projectGutenberg,manuProjectGutenberg} without further embedding adaptation are shown in Table~\ref{tab:combined-table} as the Raw column of the Mix rows. The very high scores across the board show that, with proper datasets, a model could still achieve similar performance on the two different experiments. 

However, the rightmost column in Table~\ref{tab:combined-table} also shows that embedding adaptation on the Mix dataset for models trained on other datasets could not achieve this effect -- the percentage differences between the two embedding adaptations are not sufficient to close the gap between the two experiments. 
While the previous subsections show that embedding adaptation can very much eliminate differences across natural languages, it just does not have the same ability when it comes to different experiments. Figure~\ref{fig:dataset-scatter} further shows that the models still exhibit noticeable gaps in performance between the two experiments even after embedding adaptation on the Mix dataset. Most models also exhibit a statistically significant deficit in the scores of both experiments compared with the models directly trained on the Mix dataset. As Experiment 3 includes stimuli in narrative style in addition to the Wikipedia style in Experiment 2, the results seemingly suggest that embedding adaptation is not very effective in learning across different styles, in this case, adapting to the narrative style.  
This effect might be a result of the difference in syntactic and semantic structures across domains long known to linguists \citep{biberVariationSpeechWriting1988, biberUsingRegisterDiversifiedCorpora1993}. 

\section{Conclusion}
\label{sec:discussion}

In this paper, we examine BS using a wide variety of datasets including 7 natural languages, Python, the human genome, nested parentheses (Dyck), and scrambled English. We find that language models trained on natural languages behave very similarly to each other, while other sequences lag behind this group to varying degrees, with Python performing most similarly. 
These results suggest that the factors that explain BS are not language-specific, so we should be cautious in attempting to conclude from high BS that human language processing relies on the same mechanisms as language model processing.

Methodologically, our experiments complement those of 
\citet{kaufLexicalSemanticContentNot2024}.  In a key condition, they perturb stimuli input to a pretrained English model and use the resulting representations to predict neural activations (using a regression model trained on non-perturbed stimuli).  Their results suggest that lexical semantics plays a large role in driving BS. By contrast, we manipulate the training data to the language models and then apply BS as normal. The improvement of models trained on non-natural-language data after embedding adaptation in our experiments also suggests that lexical semantics plays a non-trivial role in BS performance.
Through different angles, both studies reveal that BS is susceptible to show higher-than-expected scores with only certain properties of the English language being present in different parts of the pipeline. Beyond semantics, our experiments confirm that BS is likely to be sensitive to certain universal properties spanning across natural languages with syntax being one of the likely contributing factors. The high Python score also supports this idea as Python is semantically distinct from natural languages but shares some important syntactic features. 

The success of all natural languages---and the relatively low impact of embedding adaptation on them---can be interpreted in at least two ways.
On the one hand, this suggests that BS tracks structure that is common to all natural languages, resembling the human ability to acquire any natural language from a young age.
On the other hand, this lack of language-specificity in BS is counter-intuitive. A person reading and understanding Indonesian materials is not in the standard case processing the materials in English in their mind; yet a language model trained on Indonesian can predict neural responses of reading in English. 
Gathering cross-linguistic neural data from native speakers of many languages using the same setup as \cite{pereiraUniversalDecoderLinguistic2018} would be a promising next step.
Comparing the similarity of the responses to those of the original test pool could provide a `cross-lingual noise ceiling' comparable to the `cross-person' one used to normalize BS. 
More generally, the results from such an experiment would be indicative of how much ``language generality'' we should expect from a metric like BS. 


Future work can refine the results in this paper in several directions.  Alternative vocabulary adaptation approaches, such as Procrustes, or other non-parametric ones, could be tested as a middle ground between the raw models and embedding adaptation.  More generally, our results motivate refinements of BS, in order to come up with LM-brain similarity scores that are sensitive to the differences amongst natural languages and between natural languages and other structured sequences.  Such refinements can help deepen our understanding of the similarities and differences between language model and human language processing.

\bibliography{clmbr_brain_score,corpora}
\bibliographystyle{acl_natbib}

\end{document}

%% file: generated/language_family.tex
\begin{table}[ht]
\centering
\begin{tabular}{@{}lll@{}}
\toprule
\bf Language & \bf Family & \bf Genus \\
\midrule
Arabic & Afro-Asiatic & Semitic \\
Chinese & Sino-Tibetan & Chinese \\
English & Indo-European & Germanic \\
Indonesian & Austronesian & Malayo-Sumbawan \\
Japanese & Japanese & Japanese \\
Korean & Korean & Korean \\
Russian & Indo-European & Slavic \\
\bottomrule
\end{tabular}
\caption{Classifications of natural languages according to \citet{wals}. We consider the variety of Chinese on Wikipedia to be Mandarin, and the variety of Arabic to be Modern Standard Arabic. }
\label{tab:lang_family}
\end{table}

%% file: generated/dataset_list_table.tex
\begin{table}[t]
\centering
\begingroup
\scriptsize
\setlength{\tabcolsep}{1.0pt}
\renewcommand{\arraystretch}{1.08}
\noindent\begin{tabular}{@{}>{\centering\arraybackslash}m{0.21\columnwidth}>{\centering\arraybackslash}m{0.33\columnwidth}>{\centering\arraybackslash}m{0.40\columnwidth}@{}}
\toprule
Group & Sequence & Validation Token Count \\
\midrule
\multirow[c]{7}{0.21\columnwidth}{\centering Wikipedia} & Arabic & 40,001,536\\
& Chinese & 99,999,744\\
& English & 99,999,744\\
& Indonesian & 41,289,728\\
& Japanese & 99,999,744\\
& Korean & 48,664,576\\
& Russian & 99,999,744\\
\midrule
\multirow[c]{4}{0.21\columnwidth}{\centering Other} & Dyck & 40,001,536\\
& Genome & 99,999,744\\
& Python & 99,999,744\\
& English Mix & 99,999,744\\
\bottomrule
\end{tabular}
\endgroup
\caption{The token count for the validation set of each dataset. Scrambled English is a dataset that derives from the English Wikipedia dataset, so it has the same token count. }
\label{tab:valid}
\end{table}

%% file: generated/combined_table.tex
\begin{table*}[!tp]
\centering
\scriptsize
\setlength{\tabcolsep}{2pt}
\renewcommand{\arraystretch}{0.85}
\begin{tabular*}{\textwidth}{@{\extracolsep{\fill}}lcccccc@{}}
\toprule
 & & \multicolumn{2}{c}{Embedding Adaptation} & \multicolumn{2}{c}{$\Delta$\% Raw$\to$} & \\
\cmidrule(lr){3-4} \cmidrule(lr){5-6}
Dataset & Raw (avg $\pm$ sd) & Wikipedia (avg $\pm$ sd) & Mix (avg $\pm$ sd) & W & M & $($M$-$W$)/$W $\%$ \\
\midrule
\addlinespace[3pt]
\multicolumn{7}{@{}l}{\textbf{Experiment 2}} \\
\addlinespace[1pt]
Arabic & $0.712 \pm 0.048$ & $0.761 \pm 0.021$ & $0.740 \pm 0.026$ & \textbf{+6.9} & \textbf{+4.0} & \textbf{$-$2.7} \\
Chinese & $0.722 \pm 0.029$ & $0.772 \pm 0.032$ & $0.743 \pm 0.030$ & \textbf{+7.0} & \textbf{+3.0} & \textbf{$-$3.8} \\
English & $0.760 \pm 0.016$ & $0.760 \pm 0.016$ & $0.729 \pm 0.024$ & \textbf{$-$0.1} & \textbf{$-$4.1} & \textbf{$-$4.0} \\
Indonesian & $0.740 \pm 0.024$ & $0.770 \pm 0.025$ & $0.747 \pm 0.015$ & \textbf{+4.0} & \textbf{+1.0} & \textbf{$-$2.9} \\
Japanese & $0.729 \pm 0.031$ & $0.786 \pm 0.035$ & $0.770 \pm 0.029$ & \textbf{+7.8} & \textbf{+5.6} & \textbf{$-$2.1} \\
Korean & $0.718 \pm 0.047$ & $0.771 \pm 0.033$ & $0.762 \pm 0.046$ & \textbf{+7.4} & \textbf{+6.1} & \textbf{$-$1.2} \\
Russian & $0.712 \pm 0.035$ & $0.750 \pm 0.027$ & $0.728 \pm 0.028$ & \textbf{+5.3} & \textbf{+2.2} & \textbf{$-$2.9} \\
\addlinespace[2pt]
Dyck & $0.090 \pm 0.004$ & $0.684 \pm 0.057$ & $0.645 \pm 0.027$ & \textbf{+657.0} & \textbf{+614.0} & \textbf{$-$5.7} \\
Genome & $0.085 \pm 0.033$ & $0.720 \pm 0.023$ & $0.694 \pm 0.020$ & \textbf{+751.3} & \textbf{+721.4} & \textbf{$-$3.5} \\
Python & $0.601 \pm 0.020$ & $0.720 \pm 0.019$ & $0.708 \pm 0.018$ & \textbf{+19.7} & \textbf{+17.8} & \textbf{$-$1.6} \\
\addlinespace[2pt]
Scrambled Eng. & $0.500 \pm 0.039$ & $0.626 \pm 0.027$ & $0.621 \pm 0.020$ & \textbf{+25.2} & \textbf{+24.3} & \textbf{$-$0.7} \\
Rand.\ Init. & $0.723 \pm 0.021$ & $0.729 \pm 0.018$ & $0.714 \pm 0.028$ & \textbf{+0.8} & \textbf{$-$1.3} & \textbf{$-$2.1} \\
\addlinespace[2pt]
\textbf{Mix} & $0.804 \pm 0.028$ & $0.758 \pm 0.020$ & $0.801 \pm 0.031$ & \textbf{$-$5.7} & \textbf{$-$0.3} & \textbf{+5.7} \\
\midrule
\addlinespace[3pt]
\multicolumn{7}{@{}l}{\textbf{Experiment 3}} \\
\addlinespace[1pt]
Arabic & $0.561 \pm 0.056$ & $0.654 \pm 0.049$ & $0.625 \pm 0.042$ & \textbf{+16.6} & \textbf{+11.4} & \textbf{$-$4.4} \\
Chinese & $0.603 \pm 0.051$ & $0.633 \pm 0.041$ & $0.688 \pm 0.057$ & \textbf{+5.1} & \textbf{+14.1} & \textbf{+8.6} \\
English & $0.647 \pm 0.032$ & $0.667 \pm 0.057$ & $0.618 \pm 0.064$ & \textbf{+3.2} & \textbf{$-$4.5} & \textbf{$-$7.4} \\
Indonesian & $0.633 \pm 0.034$ & $0.648 \pm 0.018$ & $0.645 \pm 0.043$ & \textbf{+2.4} & \textbf{+1.9} & \textbf{$-$0.5} \\
Japanese & $0.606 \pm 0.033$ & $0.637 \pm 0.048$ & $0.656 \pm 0.043$ & \textbf{+5.0} & \textbf{+8.1} & \textbf{+3.0} \\
Korean & $0.578 \pm 0.051$ & $0.646 \pm 0.025$ & $0.670 \pm 0.038$ & \textbf{+11.7} & \textbf{+15.8} & \textbf{+3.7} \\
Russian & $0.631 \pm 0.061$ & $0.627 \pm 0.014$ & $0.643 \pm 0.039$ & \textbf{$-$0.7} & \textbf{+1.8} & \textbf{+2.5} \\
\addlinespace[2pt]
Dyck & $0.137 \pm 0.040$ & $0.426 \pm 0.110$ & $0.446 \pm 0.039$ & \textbf{+211.3} & \textbf{+225.9} & \textbf{+4.7} \\
Genome & $0.138 \pm 0.041$ & $0.497 \pm 0.035$ & $0.462 \pm 0.023$ & \textbf{+261.1} & \textbf{+236.1} & \textbf{$-$6.9} \\
Python & $0.485 \pm 0.016$ & $0.615 \pm 0.029$ & $0.580 \pm 0.036$ & \textbf{+26.6} & \textbf{+19.4} & \textbf{$-$5.7} \\
\addlinespace[2pt]
Scrambled Eng. & $0.243 \pm 0.034$ & $0.209 \pm 0.054$ & $0.408 \pm 0.169$ & \textbf{$-$14.0} & \textbf{+67.6} & \textbf{+94.9} \\
Rand.\ Init. & $0.228 \pm 0.031$ & $0.279 \pm 0.046$ & $0.311 \pm 0.045$ & \textbf{+22.0} & \textbf{+36.4} & \textbf{+11.8} \\
\addlinespace[2pt]
\textbf{Mix} & $0.772 \pm 0.054$ & $0.722 \pm 0.033$ & $0.764 \pm 0.048$ & \textbf{$-$6.4} & \textbf{$-$1.0} & \textbf{+5.9} \\
\bottomrule
\end{tabular*}
\caption{Effects of different training regimes. The first three columns from the left show the brain scores of models solely trained on the corresponding datasets and models with subsequent embedding adaptation on English Wikipedia data and the English mix between Project Gutenberg and Wikipedia. The third and second rightmost columns of percentage change show the comparison between raw training and embedding adaptation on Wikipedia on Mix respectively, while the rightmost column shows the differenc between the two embedding adpatations.}
\label{tab:combined-table}
\end{table*}

%% file: paper.bbl
\begin{thebibliography}{26}
\expandafter\ifx\csname natexlab\endcsname\relax\def\natexlab#1{#1}\fi

\bibitem[{Biber(1988)}]{biberVariationSpeechWriting1988}
Douglas Biber. 1988.
\newblock \emph{Variation across speech and writing}.
\newblock Cambridge University Press, Cambridge, England.

\bibitem[{Biber(1993)}]{biberUsingRegisterDiversifiedCorpora1993}
Douglas Biber. 1993.
\newblock \href {https://aclanthology.org/J93-2001/} {Using
  {Register}-{Diversified} {Corpora} for {General} {Language} {Studies}}.
\newblock \emph{Computational Linguistics}, 19(2):219--241.

\bibitem[{Cai et~al.(2024)Cai, Duan, Haslett, Wang, and
  Pickering}]{caiLargeLanguageModels2024}
Zhenguang Cai, Xufeng Duan, David Haslett, Shuqi Wang, and Martin Pickering.
  2024.
\newblock \href {https://doi.org/10.18653/v1/2024.cmcl-1.4} {Do large language
  models resemble humans in language use?}
\newblock In \emph{Proceedings of the {Workshop} on {Cognitive} {Modeling} and
  {Computational} {Linguistics}}, pages 37--56, Bangkok, Thailand. Association
  for Computational Linguistics.

\bibitem[{Chiang and Lee(2022)}]{chiangTransferabilityPretrainedLanguage2022}
Cheng-Han Chiang and Hung-yi Lee. 2022.
\newblock \href {https://doi.org/10.1609/aaai.v36i10.21295} {On the
  {Transferability} of {Pre}-trained {Language} {Models}: {A} {Study} from
  {Artificial} {Datasets}}.
\newblock \emph{Proceedings of the AAAI Conference on Artificial Intelligence},
  36(10):10518--10525.

\bibitem[{Dryer and Haspelmath(2013)}]{wals}
Matthew~S. Dryer and Martin Haspelmath, editors. 2013.
\newblock \href {https://doi.org/10.5281/zenodo.13950591} {\emph{WALS Online
  (v2020.4)}}.
\newblock Zenodo.

\bibitem[{Faysse(2023)}]{manuProjectGutenberg}
Manuel Faysse. 2023.
\newblock \href {https://huggingface.co/datasets/manu/project_gutenberg}
  {Project {Gutenberg}}.
\newblock Hugging Face Datasets.

\bibitem[{Feghhi et~al.(2024)Feghhi, Hadidi, Song, Blank, and
  Kao}]{feghhiWhatAreLarge2024}
Ebrahim Feghhi, Nima Hadidi, Bryan Song, Idan~A. Blank, and Jonathan~C. Kao.
  2024.
\newblock \href {https://doi.org/10.48550/arXiv.2406.01538} {What {Are} {Large}
  {Language} {Models} {Mapping} to in the {Brain}? {A} {Case} {Against}
  {Over}-{Reliance} on {Brain} {Scores}}.
\newblock ArXiv:2406.01538 [cs].

\bibitem[{Hosseini et~al.(2024)Hosseini, Schrimpf, Zhang, Bowman, Zaslavsky,
  and Fedorenko}]{hosseiniArtificialNeuralNetwork2024}
Eghbal~A. Hosseini, Martin Schrimpf, Yian Zhang, Samuel Bowman, Noga Zaslavsky,
  and Evelina Fedorenko. 2024.
\newblock \href {https://doi.org/10.1162/nol_a_00137} {Artificial {Neural}
  {Network} {Language} {Models} {Predict} {Human} {Brain} {Responses} to
  {Language} {Even} {After} a {Developmentally} {Realistic} {Amount} of
  {Training}}.
\newblock \emph{Neurobiology of Language}, 5(1):43--63.

\bibitem[{Hu et~al.(2025)Hu, Petty, Shi, Merrill, and
  Linzen}]{huCircuitsChomskyPrepretraining2025}
Michael~Y. Hu, Jackson Petty, Chuan Shi, William Merrill, and Tal Linzen. 2025.
\newblock \href {https://doi.org/10.18653/v1/2025.acl-long.478} {Between
  {Circuits} and {Chomsky}: {Pre}-pretraining on {Formal} {Languages} {Imparts}
  {Linguistic} {Biases}}.
\newblock In \emph{Proceedings of the 63rd {Annual} {Meeting} of the
  {Association} for {Computational} {Linguistics} ({Volume} 1: {Long}
  {Papers})}, pages 9691--9709, Vienna, Austria. Association for Computational
  Linguistics.

\bibitem[{Jiang et~al.(2026)Jiang, Shinnick, Hengel, Saratchandran, and
  Teney}]{jiangProceduralPretrainingWarming2026}
Liangze Jiang, Zachary Shinnick, Anton van~den Hengel, Hemanth Saratchandran,
  and Damien Teney. 2026.
\newblock \href {https://doi.org/10.48550/arXiv.2601.21725} {Procedural
  {Pretraining}: {Warming} {Up} {Language} {Models} with {Abstract} {Data}}.
\newblock ArXiv:2601.21725 [cs].

\bibitem[{Kauf et~al.(2024)Kauf, Tuckute, Levy, Andreas, and
  Fedorenko}]{kaufLexicalSemanticContentNot2024}
Carina Kauf, Greta Tuckute, Roger Levy, Jacob Andreas, and Evelina Fedorenko.
  2024.
\newblock \href {https://doi.org/10.1162/nol_a_00116} {Lexical-{Semantic}
  {Content}, {Not} {Syntactic} {Structure}, {Is} the {Main} {Contributor} to
  {ANN}-{Brain} {Similarity} of {fMRI} {Responses} in the {Language}
  {Network}}.
\newblock \emph{Neurobiology of Language}, 5(1):7--42.

\bibitem[{Kim et~al.(2024)Kim, Schuster, and
  Toshniwal}]{kimCodePretrainingImproves2024}
Najoung Kim, Sebastian Schuster, and Shubham Toshniwal. 2024.
\newblock \href {https://doi.org/10.48550/arXiv.2405.21068} {Code {Pretraining}
  {Improves} {Entity} {Tracking} {Abilities} of {Language} {Models}}.
\newblock ArXiv:2405.21068 [cs].

\bibitem[{Kocetkov et~al.(2022)Kocetkov, Li, Ben~Allal, Li, Mou,
  Muñoz~Ferrandis, Jernite, Mitchell, Hughes, Wolf, Bahdanau, von Werra, and
  de~Vries}]{Kocetkov2022TheStack}
Denis Kocetkov, Raymond Li, Loubna Ben~Allal, Jia Li, Chenghao Mou, Carlos
  Muñoz~Ferrandis, Yacine Jernite, Margaret Mitchell, Sean Hughes, Thomas
  Wolf, Dzmitry Bahdanau, Leandro von Werra, and Harm de~Vries. 2022.
\newblock The {Stack}: 3 {TB} of permissively licensed source code.
\newblock \emph{Preprint}.

\bibitem[{{National Center for Biotechnology
  Information}(2022)}]{NCBI_Assembly_GCF00000140540_GRCh38p14}
{National Center for Biotechnology Information}. 2022.
\newblock \href
  {https://www.ncbi.nlm.nih.gov/datasets/genome/GCF_000001405.40/} {Genome
  assembly {GRCh38.p14}}.
\newblock Accession No. GCF\_000001405.40.

\bibitem[{Papadimitriou and
  Jurafsky(2020)}]{papadimitriouLearningMusicHelps2020}
Isabel Papadimitriou and Dan Jurafsky. 2020.
\newblock \href {https://doi.org/10.18653/v1/2020.emnlp-main.554} {Learning
  {Music} {Helps} {You} {Read}: {Using} {Transfer} to {Study} {Linguistic}
  {Structure} in {Language} {Models}}.
\newblock In \emph{Proceedings of the 2020 {Conference} on {Empirical}
  {Methods} in {Natural} {Language} {Processing} ({EMNLP})}, pages 6829--6839,
  Online. Association for Computational Linguistics.

\bibitem[{Papadimitriou and
  Jurafsky(2023)}]{papadimitriouInjectingStructuralHints2023}
Isabel Papadimitriou and Dan Jurafsky. 2023.
\newblock \href {https://doi.org/10.18653/v1/2023.findings-emnlp.563}
  {Injecting structural hints: {Using} language models to study inductive
  biases in language learning}.
\newblock In \emph{Findings of the {Association} for {Computational}
  {Linguistics}: {EMNLP} 2023}, pages 8402--8413, Singapore. Association for
  Computational Linguistics.

\bibitem[{Pasquiou et~al.(2022)Pasquiou, Lakretz, Hale, Thirion, and
  Pallier}]{pasquiouNeuralLanguageModels2022}
Alexandre Pasquiou, Yair Lakretz, John~T. Hale, Bertrand Thirion, and
  Christophe Pallier. 2022.
\newblock \href {https://proceedings.mlr.press/v162/pasquiou22a.html} {Neural
  {Language} {Models} are not {Born} {Equal} to {Fit} {Brain} {Data}, but
  {Training} {Helps}}.
\newblock In \emph{Proceedings of the 39th {International} {Conference} on
  {Machine} {Learning}}, pages 17499--17516. PMLR.

\bibitem[{Pereira et~al.(2018)Pereira, Lou, Pritchett, Ritter, Gershman,
  Kanwisher, Botvinick, and Fedorenko}]{pereiraUniversalDecoderLinguistic2018}
Francisco Pereira, Bin Lou, Brianna Pritchett, Samuel Ritter, Samuel~J.
  Gershman, Nancy Kanwisher, Matthew Botvinick, and Evelina Fedorenko. 2018.
\newblock \href {https://doi.org/10.1038/s41467-018-03068-4} {Toward a
  universal decoder of linguistic meaning from brain activation}.
\newblock \emph{Nature Communications}, 9(1):963.

\bibitem[{{Project Gutenberg}(n.d.)}]{projectGutenberg}
{Project Gutenberg}. n.d.
\newblock \href {https://www.gutenberg.org/} {[link]}.

\bibitem[{Radford et~al.(2019)Radford, Wu, Child, Luan, Amodei, and
  Sutskever}]{radfordLanguageModelsAre2019}
Alec Radford, Jeffrey Wu, Rewon Child, David Luan, Dario Amodei, and Ilya
  Sutskever. 2019.
\newblock \href
  {https://cdn.openai.com/better-language-models/language_models_are_unsupervised_multitask_learners.pdf}
  {Language {Models} are {Unsupervised} {Multitask} {Learners}}.
\newblock \emph{OpenAI Blog}.

\bibitem[{Ri and Tsuruoka(2022)}]{riPretrainingArtificialLanguage2022}
Ryokan Ri and Yoshimasa Tsuruoka. 2022.
\newblock \href {https://doi.org/10.18653/v1/2022.acl-long.504} {Pretraining
  with {Artificial} {Language}: {Studying} {Transferable} {Knowledge} in
  {Language} {Models}}.
\newblock In \emph{Proceedings of the 60th {Annual} {Meeting} of the
  {Association} for {Computational} {Linguistics} ({Volume} 1: {Long}
  {Papers})}, pages 7302--7315, Dublin, Ireland. Association for Computational
  Linguistics.

\bibitem[{Schrimpf et~al.(2021)Schrimpf, Blank, Tuckute, Kauf, Hosseini,
  Kanwisher, Tenenbaum, and Fedorenko}]{schrimpfNeuralArchitectureLanguage2021}
Martin Schrimpf, Idan~Asher Blank, Greta Tuckute, Carina Kauf, Eghbal~A.
  Hosseini, Nancy Kanwisher, Joshua~B. Tenenbaum, and Evelina Fedorenko. 2021.
\newblock \href {https://doi.org/10.1073/pnas.2105646118} {The neural
  architecture of language: {Integrative} modeling converges on predictive
  processing}.
\newblock \emph{Proceedings of the National Academy of Sciences},
  118(45):e2105646118.

\bibitem[{Schrimpf et~al.(2018)Schrimpf, Kubilius, Hong, Majaj, Rajalingham,
  Issa, Kar, Bashivan, Prescott-Roy, Geiger, Schmidt, Yamins, and
  DiCarlo}]{SchrimpfKubilius2018BrainScore}
Martin Schrimpf, Jonas Kubilius, Ha~Hong, Najib~J. Majaj, Rishi Rajalingham,
  Elias~B. Issa, Kohitij Kar, Pouya Bashivan, Jonathan Prescott-Roy, Franziska
  Geiger, Kailyn Schmidt, Daniel L.~K. Yamins, and James~J. DiCarlo. 2018.
\newblock \href {https://www.biorxiv.org/content/10.1101/407007v2}
  {Brain-score: Which artificial neural network for object recognition is most
  brain-like?}
\newblock \emph{bioRxiv preprint}.

\bibitem[{Schrimpf et~al.(2020)Schrimpf, Kubilius, Lee, Murty, Ajemian, and
  DiCarlo}]{Schrimpf2020integrative}
Martin Schrimpf, Jonas Kubilius, Michael~J Lee, N~Apurva~Ratan Murty, Robert
  Ajemian, and James~J DiCarlo. 2020.
\newblock \href {https://www.cell.com/neuron/fulltext/S0896-6273(20)30605-X}
  {Integrative benchmarking to advance neurally mechanistic models of human
  intelligence}.
\newblock \emph{Neuron}.

\bibitem[{{Wikimedia Foundation}(2023)}]{wikidumps}
{Wikimedia Foundation}. 2023.
\newblock \href {https://huggingface.co/datasets/wikimedia/wikipedia}
  {Wikimedia {{Downloads}}}.

\bibitem[{Wilcox et~al.(2024)Wilcox, Futrell, and
  Levy}]{wilcoxUsingComputationalModels2024}
Ethan~Gotlieb Wilcox, Richard Futrell, and Roger Levy. 2024.
\newblock \href {https://doi.org/10.1162/ling_a_00491} {Using {Computational}
  {Models} to {Test} {Syntactic} {Learnability}}.
\newblock \emph{Linguistic Inquiry}, 55(4):805--848.

\end{thebibliography}
